\pdfoutput=1
\documentclass[10pt,twocolumn,letterpaper]{article}

\usepackage[pagenumbers]{cvpr} 

\usepackage{graphicx}
\usepackage[accsupp]{axessibility}
\usepackage{soul}
\usepackage{amsmath}
\usepackage{amssymb}
\usepackage{pifont}
\usepackage{color}
\usepackage{xcolor}
\usepackage{bm}
\usepackage{multirow}
\usepackage{diagbox}
\usepackage{soul}
\usepackage{arydshln}
\usepackage{bbm}
\usepackage{wrapfig}
\usepackage{makecell}
\usepackage{xurl}
\usepackage[bottom,flushmargin,hang,multiple]{footmisc}
\usepackage{booktabs}

\definecolor{citecolor}{HTML}{0071bc}
\usepackage[pagebackref=true,breaklinks=true,colorlinks,citecolor=citecolor,bookmarks=false]{hyperref}

\usepackage[capitalize]{cleveref}
\crefname{section}{Sec.}{Secs.}
\Crefname{section}{Section}{main}
\Crefname{table}{Table}{Tables}
\crefname{table}{Tab.}{Tabs.}
\Crefname{figure}{Figure}{Figures}
\crefname{figure}{Fig.}{Figs.}

\newcommand{\xmark}{\ding{55}}%

\def\cvprPaperID{XXXX} 
\def\confName{ICCV}
\def\confYear{2025}

\begin{document}


\title{EASG-Bench: Video Q\&A Benchmark with Egocentric Action Scene Graphs}

\author{Ivan Rodin$^{\ast\,1}$ \quad Tz-Ying Wu$^{\ast\,2}$ \quad Kyle Min$^2$ \quad Sharath Nittur Sridhar $^2$\\ \quad Antonino Furnari$^1$ \quad Subarna Tripathi$^2$ \quad Giovanni Maria Farinella$^1$ \\[1ex]
$^1$University of Catania \qquad $^2$Intel Labs\\
{\tt\footnotesize \{ivan.rodin,antonino.furnari,giovanni.farinella\}@unict.it} \\ \tt\footnotesize \{tz-ying.wu,kyle.min,sharath.nittur.sridhar,subarna.tripathi\}@intel.com
}
\maketitle


\begin{abstract}
We introduce \emph{EASG-Bench}, a question-answering benchmark for egocentric videos where question-answer pairs are created from spatio-temporally grounded dynamic scene graphs that capture actions and the relationships among the camera wearer and objects. We propose a systematic evaluation framework and evaluate several language-only and video large language models (video-LLMs) on this benchmark. We observe a performance gap between language-only and video-LLMs, especially on questions focusing on temporal ordering, thereby identifying a research gap in the area of long-context video understanding. 
To promote the reproducibility of our findings and facilitate further research, the benchmark and accompanying code are available at the following GitHub page: \url{https://github.com/fpv-iplab/EASG-bench}.

\end{abstract}

\section{Introduction}
Recent Large Language Models have been enhanced with multimodal capabilities, equipping them with the capability to chat with long-form videos ~\cite{Ren2023TimeChat,litahuang2024}, in addition to images and interleaved vision-language content.
While different question-answering benchmarks have been proposed to evaluate such abilities, they are usually generated from narrations~\cite{chen2023egoplan, cheng2024egothink}, which makes it hard to obtain grounded question-answer pairs. In contrast, the community has recently made efforts to equip egocentric videos with more structured annotations, for instance, in the form of Egocentric Action Scene Graphs (EASG), as recently proposed in~\cite{rodin2024action}.
In this paper, we propose \texttt{EASG-Bench}, a novel question-answering benchmark where question-answer pairs are generated from the EASG annotations introduced in~\cite{rodin2024action}.
Our dataset comprises over 1,807 Q\&A pairs across five categories for videos of average length of 3.1 minutes. Each question incorporates both spatial and temporal groundings, which will support grounded question-answering and language-neutral evaluations where questions are asked based on object bounding boxes rather than object categories.
We provide details on the data generation pipeline, craft an evaluation method, and benchmark different Large Language Models (LLMs) against the newly introduced benchmark.
Surprisingly, we observe that even the best video-LLM models struggle with temporal video understanding tasks compared to their pure language counterparts. 
We investigate how to bridge the gap between Video-LLM and LLM-only performances. 
We demonstrate that a chain-of-thought prompting that explicitly captures temporal ordering improves performance.
This observation suggests a future direction of research toward spatio-temporal reasoning beyond a sequence of text tokens in long-form videos.

\vspace{5pt}
\section{Related Work}
This research is related to previous investigations on benchmarks and visual question-answering with Video-LLMs.

\begin{table*}[t!]
\resizebox{\linewidth}{!}{
    \centering
    \setlength{\tabcolsep}{5pt}
    \begin{tabular}{l|ccccccccc}
    \hline
         \textbf{Benchmark} & \textbf{Untrimmed} & \textbf{Open-ended} & \textbf{Clips} & \textbf{QA Pairs} & \textbf{Categories} & \textbf{Temporally gr.} &  \textbf{Spatially gr.} & \textbf{Language-Neutral}\\
         \hline
         EgoSchema~\cite{mangalam2023egoschema} & \checkmark & \xmark & 1,981 & 5000 & - & \xmark & \xmark & \xmark \\

         EgoThink~\cite{cheng2024egothink} & \xmark & \checkmark & 595 & 700 & 6 & \xmark & \xmark & \xmark \\

         QAEgo4D~\cite{patel2025advancing} & \checkmark & \checkmark & 1,325 & 14,507 & - & \checkmark & \xmark & \xmark \\

         MultiHop-EgoQA~\cite{goletto2024amego} & \checkmark & \checkmark & 360 & 1,080 & - & \checkmark & \xmark & \xmark \\

         EgoTempo~\cite{plizzari2025omnia} & \checkmark & \checkmark & 365 & 500 & 10 & \checkmark & \xmark & \xmark \\

         AMEGO~\cite{goletto2024amego} & \checkmark & \xmark & 100 & 20,500 & 8 & \checkmark & \checkmark & \checkmark \\

    \hline
    EASG-Bench (Ours) & \checkmark & \checkmark & 221 & 1,807 & 5 &\checkmark & \checkmark & \checkmark \\
    \hline
    \end{tabular}
    }
     \caption{Comparison with existing  Egocentric Video Question Answering benchmarks.}
    \label{tab:statistics_comparison}
\end{table*}

\subsection{Egocentric Video Q\&A Benchmarks}
Several video question-answering benchmarks on egocentric videos exist in the literature, each focusing on different aspects, summarized in Table~\ref{tab:statistics_comparison}.
EgoSchema~\cite{mangalam2023egoschema} is designed to evaluate long-form video understanding through a challenging multiple-choice question-answering (MCQA) task. 
Unlike previous benchmarks that focus solely on clip length, EgoSchema introduces temporal certificate sets to quantify the intrinsic temporal reasoning required. EgoThink~\cite{cheng2024egothink} encompasses six capabilities with twelve detailed dimensions. The benchmark is constructed using selected clips from egocentric videos, with manually annotated question-answer pairs.
QAEgo4D~\cite{patel2025advancing}, based on Ego4D-NLQ, focuses on answering questions with episodic memory using a constant-size video representation. Unlike QAEgo4D, MultiHop-EgoQA~\cite{chen2025grounded} and EgoTempo~\cite{plizzari2025omnia} require reasoning across multiple segments of a video to find the correct answer. While the questions are temporally grounded in these datasets, they are not spatially grounded. AMEGO~\cite{goletto2024amego} is a recent benchmark featuring hand-object interaction tracklets, which provides spatio-temporal grounding of the questions as the proposed \texttt{EASG-Bench}. However, the Q\&As in AMEGO use a multiple-choice format rather than open-ended questions, limiting the scope of the potential responses and the ability to evaluate more complex reasoning capabilities of AI models.

\subsection{Video Question-answering with Video-LLMs}
Past investigations also proposed different approaches for video question-answering based on video-LLMs. \textbf{TimeChat}~\cite{Ren2023TimeChat} is one of the first models that follow user instructions to locate the start and end timestamps that correspond to user queries. \textbf{LITA}~\cite{litahuang2024} is framed as a temporal localization assistant that leverages time tokens to encode the time-stamps relative to the video length. \textbf{Sa2VA}~\cite{yuan2025sa2va} focuses on dense grounded understanding of videos, which combines SAM-2~\cite{ravi2024sam} with a vision-language model LLaVA~\cite{li2024llava}, and unifies text and video into a shared LLM token space. Recently, powerful video-LLM models such as \textbf{Qwen2.5-VL}~\cite{Qwen2.5-VL} that can process multimodal inputs of various sizes and extended durations have emerged.

\section{EASG-Bench}
\begin{figure*}[!t]
    \centering
\includegraphics[width=0.99\linewidth]{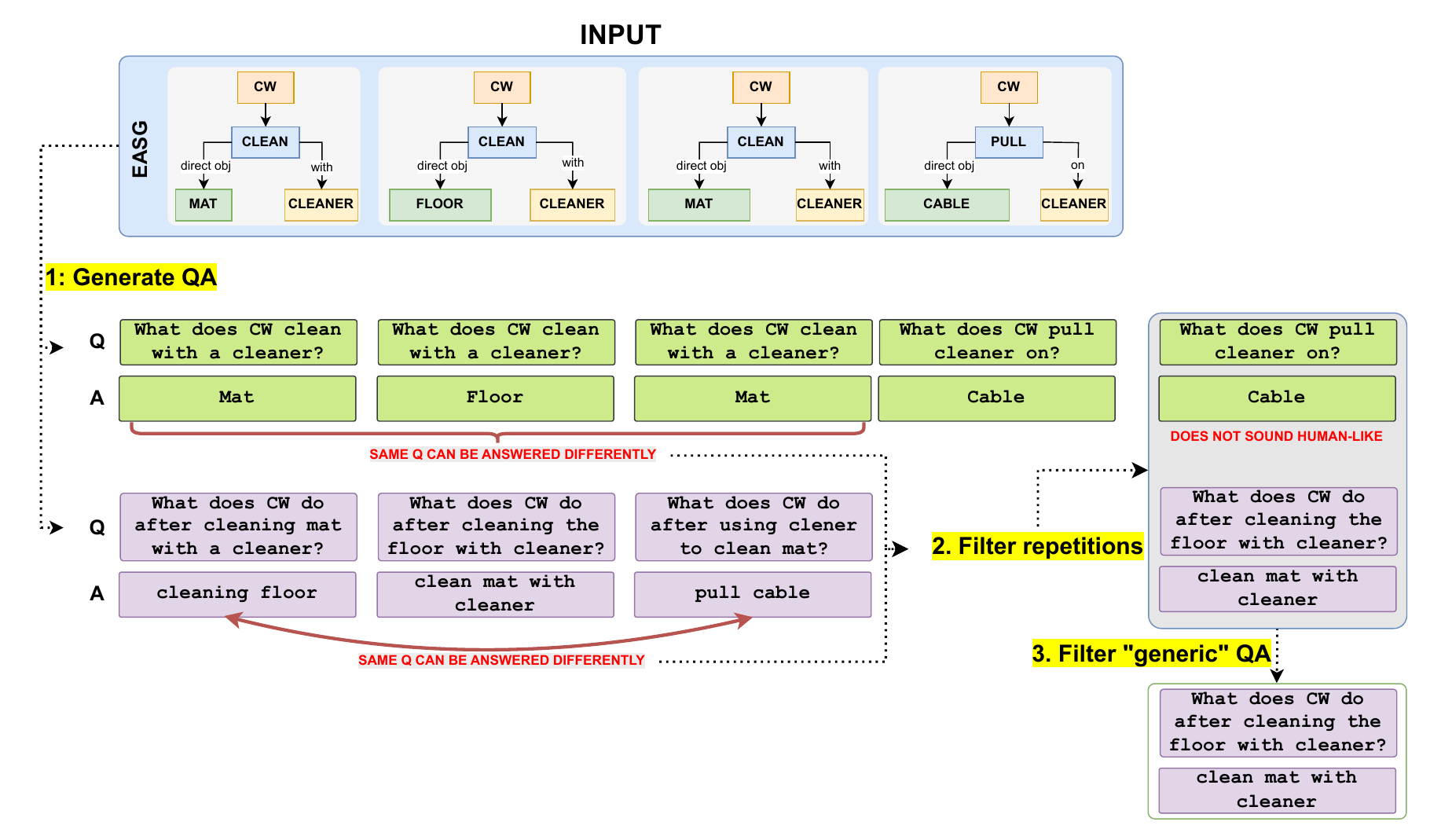}
    \caption{The 3-stage process of QA pairs generation, example for the "direct" and "after" question types. First, we prompt LLM to generate the questions from the EASG sequence, then we filter generated questions to leave only those that can be answered in unique way by observing the video clip, and finally, we filter-out the QA pairs which sound too generic.}
\label{fig:qa-generation}
\end{figure*}

\paragraph{Benchmark creation.}
We propose a novel benchmark generation approach for video question answering that leverages Egocentric Action Scene Graphs (EASGs)~\cite{rodin2024action} rather than conventional narration-based methodologies. While our work builds on the approach adopted in several previous studies, where large language models (LLMs) are employed to automatically generate question–answer pairs from narrative descriptions~\cite{plizzari2025omnia, patel2025advancing}, our method diverges by grounding the generation process in the structured representations provided by EASGs. The process of question-answer pairs generation is shown in Figure~\ref{fig:qa-generation}.

EASGs capture the intricate relationships between actors, actions, and objects within the video, enabling us to generate a categorically structured set of questions that probe various aspects of the visual content. Specifically, we systematically generate questions belonging to the following categories:

\begin{itemize}
\item \textbf{Purpose Questions}: Inquiries that explore the underlying intent of an object or an action, e.g., \textit{What is the purpose of object X in a video?}

\item \textbf{Direct Object Questions}: Questions that focus on the primary objects manipulated during an action, e.g. \textit{Q: What did camera wearer add to coffee? A: sugar}

\item \textbf{Indirect Object Questions}: Questions pertaining to secondary objects or contextual entities present during the interactions, e.g. \textit{Where did camera wearer add sugar? A: to coffee}

\item \textbf{Ordering (before/after) Questions}: Temporal queries that examine the sequence of events, such as \textit{What happened before action X?} or \textit{What happened after action X?}

\end{itemize}

\begin{figure}[t!]
    \centering
    \includegraphics[width=0.8\linewidth]{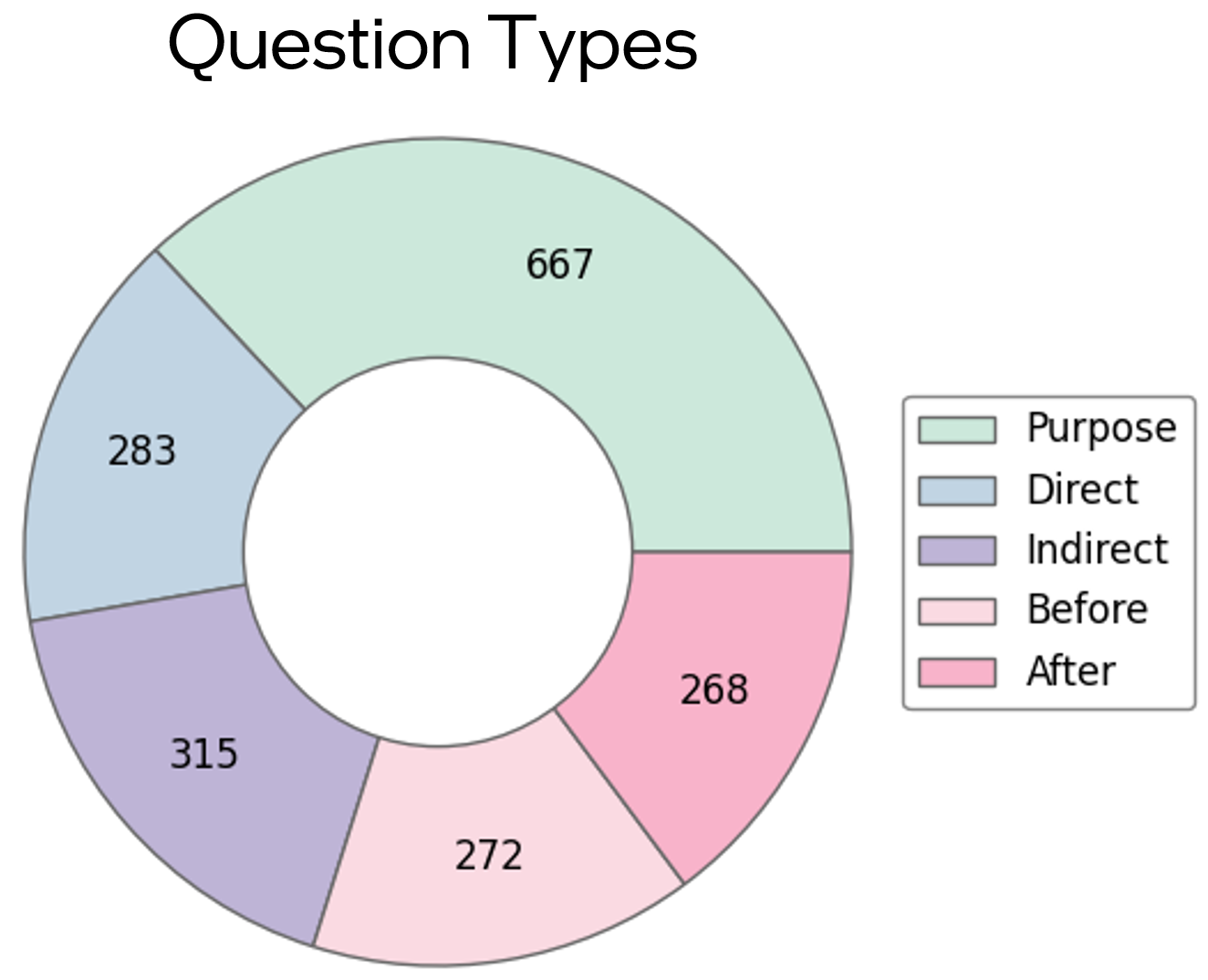}
    \vspace{-5pt}
    \caption{Dataset statistics.}
    \label{fig:stats}
\end{figure}

By anchoring the question generation process to EASGs, we ensure that each question is categorizable, structured, temporally grounded and that objects mentioned in questions and answers are spatially grounded on video frames.

To generate the question–answer pairs from the graph sequence, we condition LLMs, specifically \texttt{Llama-3.1-8B-Instruct}~\cite{grattafiori2024llama}, to restrict output to only those questions that can be unambiguously answered via direct observation of the video. This constraint is critical for long, untrimmed video benchmarking for example, given a video sequence with actions such as \textit{put cup, pour coffee, drink coffee, put cup, add sugar}, the method deliberately marks \textit{put cup} action as not fitting for the ``Ordering" QA-pair generation purpose, and avoids generating a question like ``What happened after camera wearer `put cup'?” since such an inquiry could yield multiple valid answers. To avoid hallucinations in LLM uniqueness assessment on long sequences of action scene graphs, we implement a robust filtering strategy in which we prompt Llama-3-8B with the same graph sequence extracted from a video five times. Each iteration generates an independent assessment of whether the possible QA pairs from the sequence adhere to our uniqueness criterion. We then apply majority voting across these iterations to determine the suitability of the QA pair for inclusion in our dataset.

\begin{figure*}[t!]
    \centering
    \includegraphics[width=0.85\linewidth]{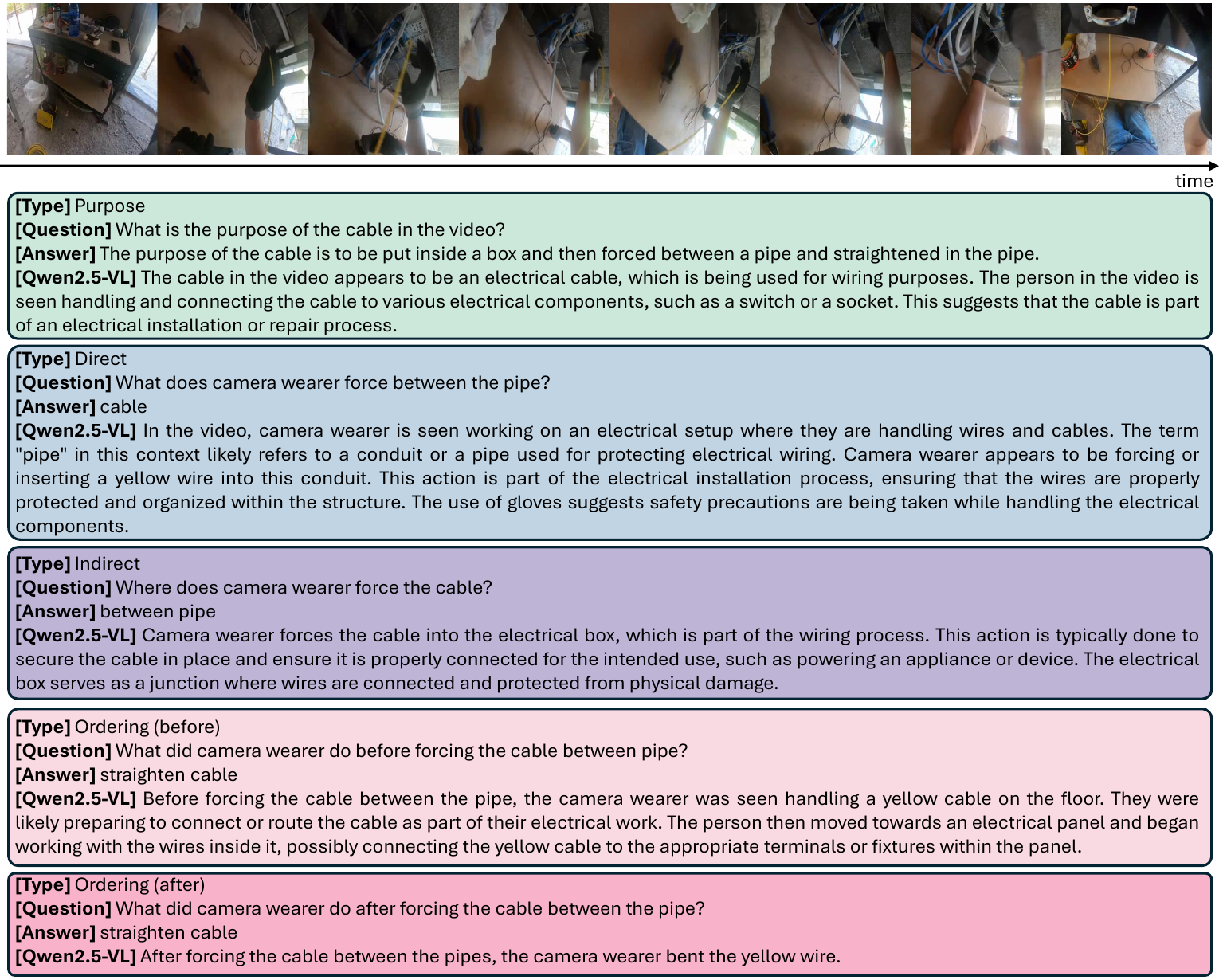}
    \vspace{-5pt}
    \caption{Samples of questions, ground truth answers, and the answer generated by Qwen2.5-VL for each type of question in \texttt{EASG-Bench}.}
    \label{fig:samples}
    \vspace{-5pt}
\end{figure*}

\vspace{-10pt}
\paragraph{Dataset statistics.}
The dataset comprises 1,807 question-answer pairs associated with 221 video clips from the Ego4D dataset. The questions are surrounded by objects grounded in EASGs. Around 1/3 of the questions are related to object functions (purpose), 1/3 about object manipulation (direct/indirect), and the remaining 1/3 about the ordering (before/after) of object manipulations. Figure~\ref{fig:stats} shows the distribution of question types, and Figure~\ref{fig:samples} shows some samples for each type of question.

\begin{figure*}[ht]
    \centering
    \includegraphics[height=0.61\linewidth,width=0.77\linewidth]{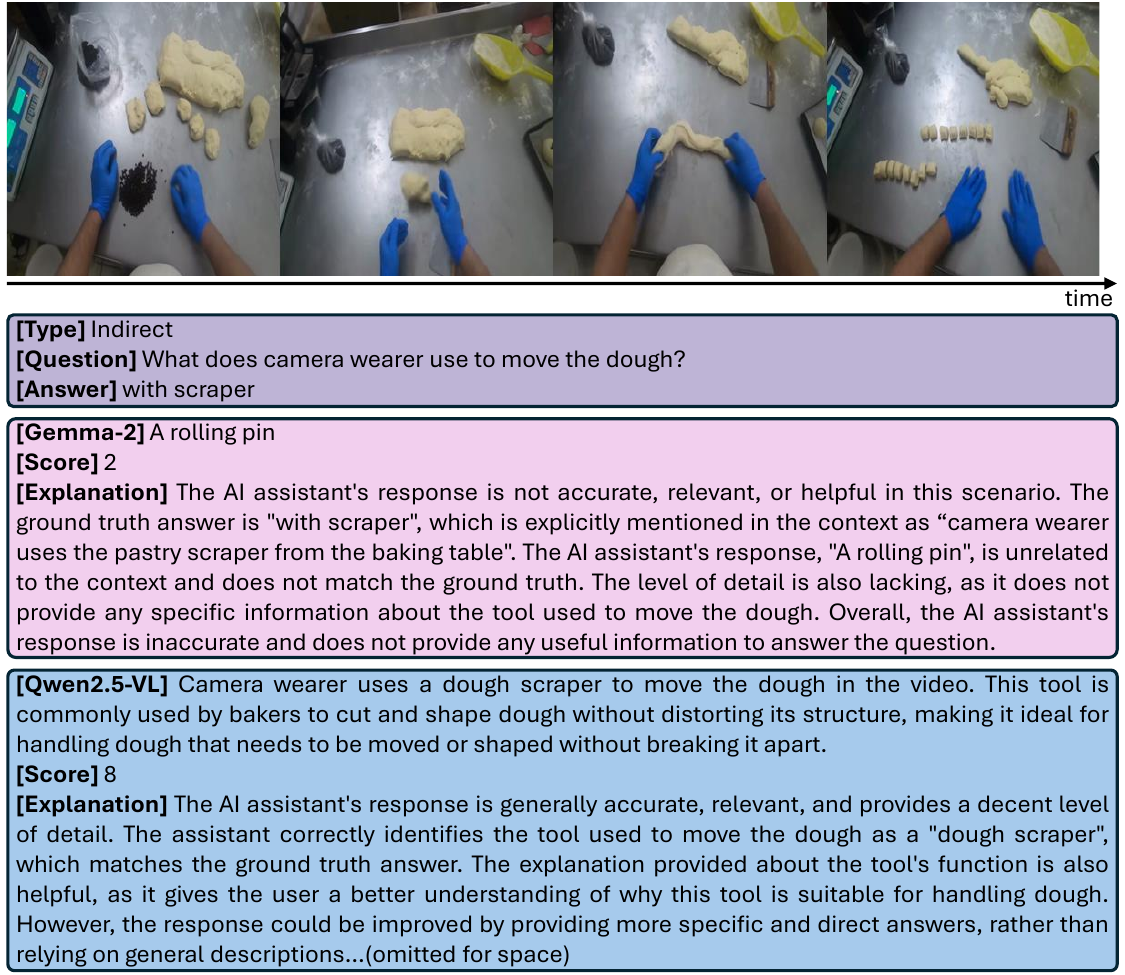}
    \vspace{-5pt}
    \caption{Examples of evaluation. The LLM judge is tasked with providing a rating to the given answer and an explanation for the rating.}
    \label{fig:eval}
    \vspace{-5pt}
\end{figure*}

\section{Experiments}
\paragraph{Evaluation.}
Our benchmark includes open-ended questions, which cannot be evaluated with standard metrics such as accuracy. Instead, we evaluate the question answering quality using {\tt Llama3-8B-Instruct}~\cite{grattafiori2024llama} with an LLM-as-a-judge mechanism similar to~\cite{litahuang2024}. 
Given the video captions with timestamps as the context information $C$ and the ground truth answer $GT$ to the question $Q$, the LLM is tasked with rating the quality of a candidate answer $A$ on a scale of 1 to 10, according to helpfulness, relevance, accuracy, and level of details.
It is also asked to provide an explanation of why the score is given to the input answer.
Specifically, the input to the LLM consists of 5 fields, i.e., ${\bf P} = [C][Q][GT][A][R]$ and the model produces a score $S$ along with an explanation $E$, where $R$ denotes the system prompt that outlines the evaluation guidelines.
Figure~\ref{fig:eval} shows an example of the evaluation.
Since the LLM is not calibrated to generate a score within the specified range, there is no guarantee that the model will give a full score to the ground truth answer. To address this, we adjust the scores by rescaling them relative to the scores given to the ground truth answer, reported as the relative scores ({\it Rel. Score}). Specifically, the relative score for a question in type $q$ is formulated as
\begin{align}
    S_i^q = \frac{LLM([C_i^q][Q_i^q][GT_i^q][A_i^q][R])}{\frac{1}{N_q}\sum_j LLM([C_j^q][Q_j^q][GT_j^q][GT_j^q][R])},
\end{align}
where $Q_i^q$ denotes the $i$-th question in the question type $q\in\{\textit{purpose},\textit{direct},\textit{indirect},\textit{before},\textit{after}\}$, $C_i^q, GT_i^q, A_i^q$ the corresponding context, ground truth and answer, respectively. $N_q$ is the total number of questions in this type.

\paragraph{Benchmark performance.}
We benchmark the proposed \texttt{EASG-Bench} with state-of-the-art (SOTA) video-LLMs, following the evaluation protocol outlined in the previous section.
Table~\ref{tab:comparison_with_baselines} presents the {\it Rel. Score} per category.
TimeChat and LITA are pioneering video-LLMs designed for tasks involving temporal understanding, such as event localization, while Sa2VA is a more recent model featuring spatial grounding with the integration of SAM2 features. Qwen2.5-VL, on the other hand, has recently been released for general-purpose video comprehension, which achieves the best performance across all the categories.
We also explore the performance of SOTA LLMs on \texttt{EASG-Bench} without providing the visual input.
While the LLMs are language-only models, they excel in answering questions about object purposes and sequences (before/after), due to their strong reasoning and common-sense capability. Video-LLMs, on the contrary, tend to be less effective for ordering-related questions, possibly because their reasoning ability is diminished when the models are tuned to accommodate a new modality.
However, for questions concerning object manipulation (direct/indirect), video-LLMs significantly outperform the language-only models, as these questions necessitate video context for an accurate response.
Nonetheless, \texttt{EASG-Bench} is still challenging for all the existing models.

\begin{table}[t!]
    \centering
    \setlength{\tabcolsep}{4pt}
    \resizebox{\linewidth}{!}{
    \begin{tabular}{lcccccc} 
        \toprule
        Models & Purpose & Direct & Indirect & Before & After & Avg. \\ 
         \cmidrule(lr){1-1} \cmidrule(lr){2-7}
        \multicolumn{7}{l}{\textit{Language-only}} \\
        Gemma-2-9B~\cite{team2024gemma}
          & 57.98 & 37.34 & 50.13 & 94.08 & 60.48 & 60.00 \\
        Llama-3.1-8B-Instruct~\cite{grattafiori2024llama}
          & 64.67 & 39.11 & 45.60 & 92.46 & 67.74 & 61.92 \\
        Qwen3-8B~\cite{yang2025qwen3}
          & 69.37 &	55.06 &	56.22 &	\textbf{98.92} &	\textbf{70.81} &	70.08 \\
        \cmidrule(lr){1-1} \cmidrule(lr){2-7}
        \multicolumn{7}{l}{\textit{Video-LLM}} \\
        TimeChat-7B~\cite{Ren2023TimeChat}
          & 47.44	& 58.23	& 55.31 & 61.22 & 49.52 & 54.34 \\
        LITA-13B~\cite{litahuang2024}
          & 50.14	& 40.51 & 43.26 & 53.68 & 40.32 & 45.58 \\
        Sa2VA-8B~\cite{yuan2025sa2va}
          & 55.98	& 57.09 & 65.16 & 67.50 & 53.55 & 59.86 \\
        Qwen2.5-VL-7B~\cite{Qwen2.5-VL}
          & \textbf{75.78} & \textbf{70.38} & \textbf{72.02} & 82.76 & 65.97 & \textbf{73.38} \\
        \bottomrule
    \end{tabular}
    }    
    \captionof{table}{
Benchmarking results with LLMs and video-LLMs.
}\label{tab:comparison_with_baselines}
\end{table}

\begin{table}[t!]
    \centering
    {
    \setlength{\tabcolsep}{5pt}
    \resizebox{0.9\linewidth}{!}{
    \begin{tabular}{lccc}
        \toprule
         Prompting & Before & After & Avg  \\
         \cmidrule(lr){1-1} \cmidrule(lr){2-4}
         Regular  & 82.76 & 65.97 & 74.37 \\
         CoT & 88.33 $(\uparrow 5.57)$ & 69.03 $(\uparrow 3.06)$ & 78.68 $(\uparrow 4.32)$ \\
         \bottomrule
    \end{tabular}
    }
    }
    \vspace{-5pt}
    \captionof{table}{Effect of Chain-of-Thought prompting on temporal order questions (i.e., ``before'' and ``after'' type) with Qwen2.5-VL-7B.}
    \label{tab:two_stage}
\end{table}

\paragraph{Chain-of-Thought prompting.}
To investigate why the video-LLMs struggle with sequence-related (before/after) questions, we delve into the idea of chain-of-thought prompting with the best-performing model, Qwen2.5-VL.
Instead of directly asking ``What did camera wearer do before/after action X?", we can prompt the model for locating action X in the first stage, and inquire about the action before/after that step in the second.
Table~\ref{tab:two_stage} shows a notable gain of this 2-stage prompting strategy compared to the 1-stage baseline. This strategy largely mitigates the gap between the video-LLMs and LLMs for ``after" questions. However, it still underperforms LLMs for ``before" questions, which suggests that there is still room for video-LLMs to improve their {\it look-back} ability.

\section{Conclusions}
We introduced \texttt{EASG-Bench}, a novel egocentric video question-answering benchmark grounded in space and time.  
A total of 1,807 questions incorporate both spatial and temporal groundings and are divided into five different categories. We evaluate a range of language-only and video-LLM models on this benchmark.
Notably, models like Qwen2.5-VL demonstrate effective use of visual signals and consistently outperform language-only baselines across most question types. However, they struggle with temporal comprehension tasks, particularly those involving reasoning over events occurring before or after.
To better understand this discrepancy, we explore the performance gap in temporal reasoning between language-only and video-LLM models,
and observe that chain-of-thought prompting can bridge the gap to some extent. The results highlight the necessity of  
future research on spatio-temporal reasoning that goes beyond textual token sequence, especially in the context of long-form video understanding.

\section*{Acknowledgments}
This research is supported by Intel Corporation. Research at the University of Catania is supported in part by the project Future Artificial Intelligence Research (FAIR) – PNRR MUR Cod. PE0000013 - CUP: E63C22001940006.

{\small
\bibliographystyle{ieee_fullname}

}

\end{document}